\documentclass[10pt,twocolumn,letterpaper]{article}

\usepackage{iccv}              % To produce the CAMERA-READY version
\usepackage[accsupp]{axessibility}
\usepackage{ragged2e} 
\usepackage{booktabs}
\usepackage{multirow}
\usepackage{adjustbox}
\usepackage[normalem]{ulem}
\definecolor{iccvblue}{rgb}{0.21,0.49,0.74}
\usepackage[pagebackref,breaklinks,colorlinks,allcolors=iccvblue]{hyperref}

\def\paperID{*****} % *** Enter the Paper ID here
\def\confName{ICCV}
\def\confYear{2025}

\title{Task-Specific Dual-Model Framework for Comprehensive Traffic Safety Video Description and Analysis}

\author{
Blessing A. Kyem$^{1}$\thanks{Equal contribution} \quad
Neema J. Owor$^{2}$\footnotemark[1] \quad
Andrews Danyo$^{1}$\footnotemark[1] \quad
Joshua K. Asamoah$^{1}$ \quad
Eugene Denteh$^{1}$ \\
Tanner Muturi$^{2}$ \quad
Anthony Dontoh$^{3}$ \quad
Yaw Adu-Gyamfi$^{2}$ \quad
Armstrong Aboah$^{1}$\thanks{Corresponding author} \\
\vspace{0.5em}
$^1$North Dakota State University \quad
$^2$University of Missouri–Columbia \quad
$^3$University of Memphis\\
\vspace{0.3em}
\parbox{\textwidth}{\centering\footnotesize%
\texttt{\{blessing.kyem,joshua.asamoah, andrews.danyo, eugene.denteh, armstrong.aboah\}@ndsu.edu}\\
\texttt{\{nodyv, twmtyg, adugyamfiy\}@missouri.edu} \\
\texttt{adontoh@memphis.edu}%
}\\
\vspace{0.3em}
\small{$^*$First Authors} \quad
\small{$^\dagger$Corresponding author}
}

\vspace{-0.5em}
\begin{document}
\maketitle
\begin{abstract}
Traffic safety analysis requires complex video understanding to capture fine-grained behavioral patterns and generate comprehensive descriptions for accident prevention. In this work, we present a unique dual-model framework that strategically utilizes the complementary strengths of VideoLLaMA and Qwen2.5-VL through task-specific optimization to address this issue. The core insight behind our approach is that separating training for captioning and visual question answering (VQA) tasks minimizes task interference and allows each model to specialize more effectively. Experimental results demonstrate that VideoLLaMA is particularly effective in temporal reasoning, achieving a CIDEr score of 1.1001, while Qwen2.5-VL excels in visual understanding with a VQA accuracy of 60.80\%. Through extensive experiments on the WTS dataset, our method achieves an S2 score of 45.7572 in the 2025 AI City Challenge Track 2, placing 10th on the challenge leaderboard. Ablation studies validate that our separate training strategy outperforms joint training by 8.6\% in VQA accuracy while maintaining captioning quality.
\end{abstract}

\vspace{-1em}    
\section{Introduction}
\label{sec:intro}

\begin{figure}
    
    \centering
    \includegraphics[width=8cm]{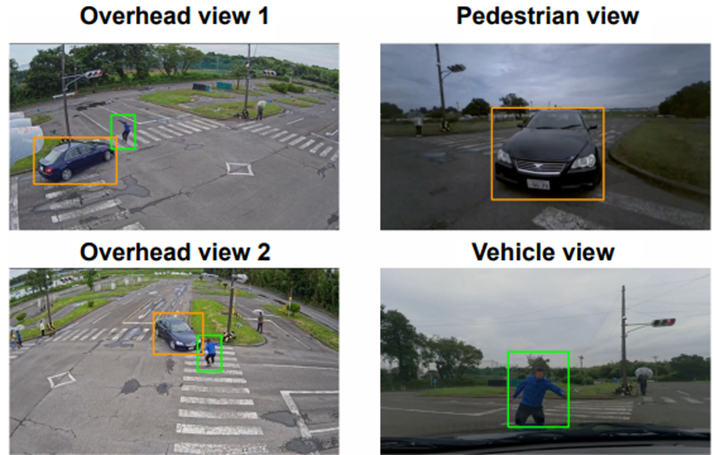}
    \begingroup
    \justifying
    \fontsize{7}{8.5}\selectfont

\noindent\textbf{Caption pedestrian:} 
    \textcolor{red}{The pedestrian, a male in his 20s and about 170 cm tall, \uline{ was walking across a zebra-striped crosswalk situated at the centre of a four-way urban intersection}.} 
    \textcolor{red}{He wore a vivid blue hooded jacket and dark trousers} 
    \textcolor{cyan}{The weather was overcast yet bright, with dry, level asphalt and no standing water.} 
    \textcolor{blue}{His body was oriented perpendicular to the vehicle’s path, moving from the driver’s right toward the opposite curb.} 
    \textcolor{orange}{Although he cast a brief glance toward the roadway, he seemed mainly focused on reaching the other side, showing little reaction to the approaching sedan.} 
    \textcolor{blue}{The relative distance between pedestrian and vehicle was moderate and decreasing, while surrounding traffic remained light. It was a weekday, and the pedestrian seemed to be aware of the approaching vehicle.}
    
    \vspace{0.3em}
    
    % Caption: Vehicle
    \noindent\textbf{Caption vehicle:} 
    \textcolor{green}{The vehicle was moving at a constant speed of 10km/h.} 
    \textcolor{blue}{It trailed the pedestrian by few metres and remained diagonal to the crosswalk, slightly to the pedestrian’s left.} 
    The vehicle had a clear view of the pedestrian. 
    \textcolor{green}{It was going diagonally and towards the direction of the pedestrian.} 
    \textcolor{red}{The environment conditions indicated that the pedestrian was a male in his 20s with a height of 170 cm. He was wearing a blue hooded jacket and dark trousers.} 
    \textcolor{blue}{The event took place in an urban area on a weekday. The weather was overcast, but the brightness was bright. The road surface was dry and level, made of asphalt. The traffic volume was very low on all the lanes on the road. Sidewalks were present on both sides of the road.}
    \vspace{0.3em}
    
    \caption{Multi-view sample from the WTS caption dataset: Overhead Views 1 \& 2 (left), Pedestrian View (top-right), and Vehicle View (bottom-right).}
    \label{fig:fish}
    \vspace{-1.7em}
    \endgroup
\end{figure}
\vspace{-1em}
Fine-grained video captioning of traffic safety scenarios has emerged as a transformative approach for understanding and preventing pedestrian accidents, enabling detailed analysis of behavioral patterns, environmental conditions, and contextual factors that traditional detection systems fail to capture \cite{Du_article, lana2021data}. The complexity of generating comprehensive textual descriptions from multi-perspective traffic footage presents unique challenges in computer vision, requiring advanced techniques that can synthesize temporal dynamics, spatial relationships, and causal narratives across continuous video sequences  \cite{DU2025}. Currently deployed ITS predominantly rely on traditional post-incident reconstruction methods and binary classification systems, offering limited descriptive capabilities in complex pedestrian accident scenarios \cite{SCHUBERT2023107139,11010810}. Comprehensive analysis systems require enhanced coverage and a detailed understanding of behavioral patterns to overcome these limitations \cite{RASHID2024100277}. However, these conventional approaches often fail to deliver continuous and detailed representations of the monitored scenarios, which remains a fundamental obstacle in developing proactive safety management systems.

Advanced video understanding now employs long fine-grained captioning to analyze traffic behaviors, particularly in vehicle-pedestrian accident scenarios \cite{dinh2024trafficvlm,10839383} (see Fig.~\ref{fig:fish}). This method facilitates continuous monitoring and detailed description of complex traffic events throughout their temporal progression using synchronized footage from multiple cameras and viewpoints, including static overhead cameras and moving vehicle ego cameras \cite{shoman2024enhancing} as illustrated in Fig.~\ref{fig:fish}. This includes TrafficVLM \cite{dinh2024trafficvlm} which analyzes traffic video events at various spatial and temporal resolutions, generating extended, detailed descriptions covering both vehicle and pedestrian activities across different phases of an incident \cite{zhou2024vision, onsu2025leveragingmultimodalllmsassistedinstance}. This enhanced analytical capability significantly advances traditional binary classification systems by providing rich temporal narratives that capture behavioral patterns, environmental conditions, and evolving risk factors integral to incident development \cite{MAHBOUBI2024104004,ALBAHRI2024109409}.

Current video captioning methods face critical limitations in multi-perspective traffic safety scenarios \cite{to2024multiperspective}. VideoLLaMA \cite{Zhang2025VideoLLaMA3F} and Video-LLaVA \cite{Lin2023VideoLLaVALU} excel at long context understanding and temporal modeling, generating comprehensive captions across extended sequences \cite{inproceedingszhang}, but suffer from hallucination artifacts and data efficiency constraints \cite{ma2024vista}. Conversely, Qwen-VL \cite{Bai2023QwenVLAV} and InternVL \cite{chen2024internvl} demonstrate superior visual reasoning and state-of-the-art VQA performance through robust spatial understanding \cite{bai2023qwenvlversatilevisionlanguagemodel}, yet struggle with extended temporal contexts and computational scalability \cite{qwen2024qwen2}. Both architectures face synchronization overhead in multi-camera deployments, limiting real-time applications. Integrating Video-LLaMA's temporal reasoning with Qwen-VL's visual reasoning presents a complementary solution where each model's strengths address the other's limitations.

To address these challenges, we propose a novel approach combining VideoLLaMA and QwenVL for traffic safety analysis. Our method utilizes VideoLLaMA's temporal reasoning and narrative generation with QwenVL's visual reasoning and spatial analysis. Our contributions have been summarized below:

\begin{itemize}
 \item We propose an approach that systematically integrates VideoLLaMA's temporal reasoning with Qwen-VL's visual reasoning capabilities for traffic safety analysis. 

 \item We empirically validate that separate task-specific training for captioning and visual question answering significantly outperforms joint training approaches.

 \item We conduct extensive experiments on the WTS dataset across multiple metrics, placing 10th in the AI City 2025 Challenge Track 2.
    
\end{itemize}

% To address this critical gap, this project will develop a multi-perspective, long-duration, fine-grained video captioning system specifically tailored for traffic safety analysis, with a particular emphasis on pedestrian-involved incidents. The proposed system addresses key analytical tasks, including …, using the Woven Traffic Safety (WTS), which provides synchronized multi-camera footage along with detailed annotations of pedestrian behavior and gaze. This integrated approach represents a significant advancement toward comprehensive traffic safety understanding that bridges the gap between traditional detection methods and the sophisticated narrative analysis required for effective safety management.

%  Our study proposes ... 

%-------------------------------------------------------------------------

\section{Related Work}
\label{sec:review}

\subsection{Vision Language Models for Traffic Captioning and Reasoning}
The past two years have seen specialized VLMs such as TrafficVLM \cite{dinh2024trafficvlm}, Wolf \cite{Li_2024_Wolf} and VLM-Auto \cite{Guo_2024_VLMAuto} set new standards in traffic video captioning by combining multi-task fine-tuning, mixture-of-experts backbones and simulator-integrated pipelines.  At the same time, general models such as VideoLLaMA \cite{damonlpsg2023videollama, damonlpsg2024videollama2, Zhang2025VideoLLaMA3F}, InternVL \cite{chen2024internvl} and Qwen-VL \cite{Bai2023QwenVLAV} have been effectively adapted via domain fine-tuning and LoRA to achieve up to 60\% VQA accuracy in pedestrian-vehicle reasoning \cite{DriveLLaVa2024}.  Dedicated VQA benchmarks including NuScenes-QA \cite{qian2023nuscenes}, LingoQA \cite{marcu2023lingoqa} and DriveLM \cite{Sima2024ECCV} have further advanced traffic-specific visual reasoning through graph-based and free-form QA formats, exposing gaps such as GPT-4V’s 59.6 \% truthfulness score \cite{marcu2023lingoqa,MuturiAIC25,OworAIC25}.  

\subsection{Architectural Advances, Benchmarks, and Deployment Challenges}
Domain-specialized architectures such as EMMA \cite{Hwang_2024_EMMA}, DriveVLM \cite{DriveVLM2024} and MAPLM \cite{MAPLM2024} utilize end-to-end multimodal fusion, chain-of-thought planning and HD-map integration to tackle safety-critical reasoning and spatiotemporal modeling.  Hybrid training strategies such as joint multi task learning \cite{Zhang_2023_Musketeer}, supervised with contrastive learning \cite{Radford_2021_CLIP, Xie_2023_RA-CLIP}, and LoRA fine tuning \cite{Hu_2021_LoRA} consistently outperform separate pipelines \cite{Shi_2024_ScVLM}. Benchmarks like WTS \cite{kong2024} and the AI City Challenge competition \cite{Wang2024AI8th} offer rigorous evaluation of caption quality, VQA accuracy, and event phase understanding.  However, persistent challenges remain in hallucination mitigation, balancing spatial resolution against temporal coverage, and achieving real-time deployment on edge hardware prompting solutions like Visual–Textual Intervention and encoder ensembles \cite{HallucinationMitigation2025}.  

\vspace{-0.5em}
\section{Proposed Method}
\begin{figure*}
  \centering
  % Adjust width (or height) as needed:
  \includegraphics[width=0.7\textwidth]{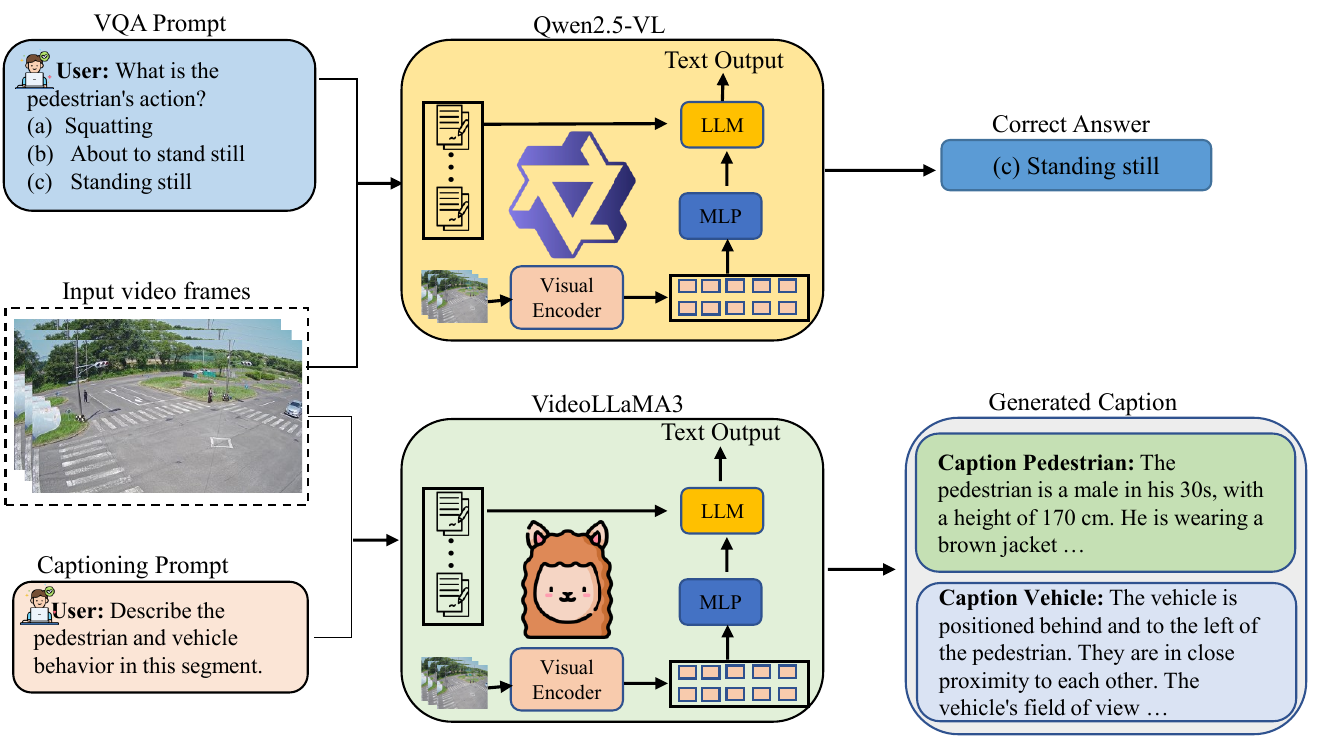}
  \caption{Overview of our proposed dual-task optimization framework.}
  \label{fig:framework}
\end{figure*}

Figure \ref{fig:framework} illustrates our proposed dual-task optimisation framework, which uses two specialized models for comprehensive traffic safety analysis. The framework processes input video frames through two parallel pathways: Qwen2.5-VL handles Visual Question Answering (VQA) tasks. Each model is independently optimized for its specific task using Low-Rank Adaptation (LoRA), preventing task interference while maximizing individual model performance.

\subsection{Problem Formulation}
Given a video $\mathcal{V} = \{v_t\}_{t=1}^T$ with $T$ frames segmented into $N$ behavioral phases $\mathcal{S} = \{s_i\}_{i=1}^N$, we address two complementary tasks: (1) generating fine-grained captions $\mathcal{C} = \{c_p^i, c_v^i\}_{i=1}^N$ describing pedestrian ($p$) and vehicle ($v$) behaviors, and (2) answering $M$ safety-critical questions $\mathcal{Q} = \{q_j\}_{j=1}^M$ with corresponding answers $\mathcal{A} = \{a_j\}_{j=1}^M$.

\subsection{Dual-Task Optimization Framework}
We employ VideoLLaMA3 with parameters $\theta$ (denoted as $f_\theta$) and Qwen2.5-VL with parameters $\phi$ (denoted as $g_\phi$) enhanced with task-specific Low-Rank Adaptation (LoRA). For each model's weight matrix $W \in \mathbb{R}^{d \times d}$, we define separate LoRA decompositions:

% \begin{figure*}
%   \centering
%   % Adjust width (or height) as needed:
%   \includegraphics[width=0.7\textwidth]{ICCV2025-Author-Kit/sec/overall framework.pdf}
%   \caption{Overview of our proposed dual-task optimization framework.}
%   \label{fig:framework}
% \end{figure*}

\begin{equation}
\begin{aligned}
W' &= W + \Delta W, \\
\text{where} \quad \Delta W &= BA, \quad B \in \mathbb{R}^{d \times r}, A \in \mathbb{R}^{r \times d}
\end{aligned}
\end{equation}

with rank $r \ll d$, where $d$ is the model dimension. We optimize distinct parameters $\{B_{cap}, A_{cap}\}$ for captioning and $\{B_{vqa}, A_{vqa}\}$ for VQA.

\textbf{Captioning Objective.} For each segment $s_i$, we minimize the negative log-likelihood of generating captions:

\begin{equation}
\mathcal{L}_{cap} = -\sum_{i=1}^{N} \sum_{k \in \{p,v\}} \sum_{t=1}^{L_k^i} \log P(c_k^{i,t} | c_k^{i,<t}, s_i; \theta, B_{cap}, A_{cap})
\end{equation}

where $c_k^{i,t}$ denotes the $t$-th token of caption type $k \in \{p, v\}$ for segment $i$, $c_k^{i,<t} = \{c_k^{i,1}, ..., c_k^{i,t-1}\}$ represents preceding tokens, $L_k^i$ is the length of caption $k$ for segment $i$, and $P$ denotes the probability distribution over the vocabulary.

\textbf{VQA Objective.} Given question-answer pairs, we optimize:

\begin{equation}
\mathcal{L}_{vqa} = -\sum_{j=1}^{M} \log P(a_j | q_j, s_j; \phi, B_{vqa}, A_{vqa})
\end{equation}

where $a_j \in \{1, ..., K\}$ represents the correct answer among $K$ multiple-choice options for question $q_j$ on segment $s_j$.

\subsection{Training Strategy}
We adopt separate training to prevent task interference:

\begin{equation}
\begin{aligned}
\{\theta^*, B_{cap}^*, A_{cap}^*\} &= \arg\min_{\theta, B_{cap}, A_{cap}} \mathcal{L}_{cap} \\
\{\phi^*, B_{vqa}^*, A_{vqa}^*\} &= \arg\min_{\phi, B_{vqa}, A_{vqa}} \mathcal{L}_{vqa}
\end{aligned}
\end{equation}

where $*$ denotes optimal parameters. This decoupled optimization preserves each model's specialized capabilities.

% \subsection{Inference}
% During inference, for a test segment $s_{test}$:

% \textbf{Caption Generation:} We employ beam search with beam width $b$ to generate caption $\hat{c}_k$:
% \begin{equation*}
% \hat{c}_k = \arg\max_{c} P(c | s_{test}; \theta^*, B_{cap}^*, A_{cap}^*)
% \end{equation*}

% \textbf{VQA:} For question $q$, we select the predicted answer $\hat{a}$:
% \begin{equation}
% \hat{a} = \arg\max_{a \in \{1,...,K\}} P(a | q, s_{test}; \phi^*, B_{vqa}^*, A_{vqa}^*)
% \end{equation}

% The complementary outputs provide a comprehensive understanding of the traffic scene.

% \vspace{-1.2em}

\section{Experiment}
\subsection{Datasets}
Our model is trained and evaluated using the WTS dataset \cite{kong2024wts} for Track 2 of the AI City Challenge 2025. The dataset contains 810 internal videos (155 scenarios) and 3,400 external pedestrian-related clips from BDD100K, representing over 130 staged traffic scenarios captured at 1080p/30fps using vehicle-mounted and overhead cameras. Each video is segmented into five behavioral phases (pre-recognition, recognition, judgment, action, avoidance) with dual-perspective captions (pedestrian and vehicle viewpoints). The dataset includes manually annotated bounding boxes with object tracking, 3D pedestrian gaze vectors, and a Traffic VQA dataset with 180 structured questions for spatial-temporal understanding.

\subsection{Evaluation Metrics}
We evaluate our approach using the official metrics established for the AI City Challenge Track 2 2025: Traffic Safety Description and Analysis. The evaluation framework employs a dual-task assessment methodology that comprehensively measures both natural language generation quality and visual reasoning capabilities through two complementary sub-tasks.

\subsubsection{Caption Generation}
The caption generation component evaluates the quality of automatically generated natural language descriptions for traffic safety scenarios. For each video segment, models must produce two detailed captions: one describing pedestrian behavior and one describing vehicle behavior throughout the temporal sequence. Performance is assessed using a composite metric that averages four established natural language generation measures: BLEU-4, METEOR, ROUGE-L, and CIDEr.

\textbf{BLEU-4} evaluates n-gram precision between generated and reference captions, defined as:
\begin{equation}
\text{BLEU-4} = BP \cdot \exp\left(\sum_{n=1}^{4} w_n \log p_n\right)
\end{equation}
where $p_n$ is the modified n-gram precision, $w_n = 1/4$ for uniform weighting, and $BP$ is the brevity penalty to discourage overly short generations.

\textbf{METEOR} evaluates translation quality by combining precision-recall harmonic mean with a penalty term that accounts for word order fragmentation:

\begin{equation}
\text{METEOR} = F_{\text{mean}} \times (1 - \text{Penalty})
\end{equation}

The harmonic mean of precision and recall is computed as:

\begin{equation}
F_{\text{mean}} = \frac{P \cdot R}{\alpha \cdot P + (1-\alpha) \cdot R}
\end{equation}

where precision $P$ and recall $R$ are defined in terms of matched unigrams:

\begin{equation}
P = \frac{m}{w_t}, \quad R = \frac{m}{w_r}
\end{equation}

The fragmentation penalty is calculated based on the number of contiguous chunks:

\begin{equation}
\text{Penalty} = \gamma \cdot \left(\frac{ch}{m}\right)^{\beta}
\end{equation}

Here, $m$ represents matched unigrams between hypothesis and reference, $w_t$ and $w_r$ denote the total unigrams in hypothesis and reference respectively, $ch$ indicates the number of contiguous matched chunks, with standard parameter values $\alpha = 0.9$, $\beta = 3$, and $\gamma = 0.5$.

\textbf{ROUGE-L} evaluates caption quality by measuring sequence-level similarity through longest common subsequences, which is particularly relevant for traffic safety descriptions where maintaining temporal coherence is crucial. The metric is formulated as
\begin{equation}
ROUGE-L = \frac{(1 + \beta^2) \times R_{lcs} \times P_{lcs}}{R_{lcs} + \beta^2 \times P_{lcs}}
\end{equation}

where:

\begin{equation}
R_{lcs} = \frac{LCS(X,Y)}{m}, P_{lcs} = \frac{LCS(X,Y)}{n}
\end{equation}

where $LCS(X,Y)$ is the length of longest common subsequence between candidate $X$ and reference $Y$, $m$ is the length of reference sequence, $n$ is the length of candidate sequence, and $\beta = P_{lcs}/R_{lcs}$ when $R_{lcs} \neq 0$.

\textbf{CIDEr} measures the consensus between generated captions and human references by emphasizing n-grams that are distinctive and informative, making it suitable for evaluating detailed traffic safety descriptions that require specific terminology and behavioral descriptors.

\begin{equation}
CIDEr_n(c_i, S_i) = \frac{1}{m} \sum_{j} \frac{g^n(c_i) \cdot g^n(s_{ij})}{|g^n(c_i)| |g^n(s_{ij})|}
\end{equation}

% where:

% \begin{equation}
% g_k^n(s) = tf_k(s) \cdot idf_k
% \end{equation}

% where $tf_k(s) = \frac{h_k(s)}{\sum_{w_l \in \Omega} h_l(s)}$ represents the term frequency of n-gram $w_k$ in sentence $s$ and $tf_k(s) = \frac{h_k(s)}{\sum_{w_l \in \Omega} h_l(s)}$ represents the term frequency of n-gram $w_k$ in sentence $s$, and $idf_k = \log \left(\frac{|I|}{\sum_{I_p \in I} \min(1, \sum_{s_q \in S_{I_p}} h_k(s_q))}\right)$ represents the inverse document frequency.

\begin{equation}
CIDEr(c_i, S_i) = \sum_{n=1}^{N} \frac{1}{N} CIDEr_n(c_i, S_i)
\end{equation}

where $g_n(s)$ represents the weighted word combinations for caption $s$, $m$ is the number of reference captions, and $N$ is the maximum word combination length.

\subsubsection{Visual Question Answering}
The second sub-task measures the model’s capability to reason visually and provide accurate answers to safety-critical questions. The performance metric employed is simple top-1 accuracy:
\begin{equation}
\mathrm{Acc} = \frac{1}{N}\sum_{i=1}^{N}\mathbf{1}\bigl[\hat{a}_i = a_i^{\star}\bigr]
\end{equation}
where \(N\) is the total number of questions, \(\hat{a}_i\) is the predicted answer, \(a_i^{\star}\) the ground-truth answer, and \(\mathbf{1}[\cdot]\) denotes the indicator function.

The overall performance is summarized into a composite final metric (\(S2\)), calculated by combining the caption generation (Cap\_Score) and VQA accuracy (Acc) metrics  as follows:

\begin{equation}
\mathrm{Cap\_Score} = \frac{\text{BLEU-4} + \text{METEOR} + \text{ROUGE-L} + \text{CIDEr}}{4}
\end{equation}
\begin{equation}
\mathrm{S2} = \frac{\mathrm{Cap\_Score} + \mathrm{Acc}}{2}
\end{equation}

% \subsection{Implementation Details}
% We implement our experiments using two state-of-the-art Video-Large Language Models: QwenVL-2.5 and VideoLLaMA.
% For QwenVL-2.5, training is conducted with a batch size of 4 and gradient accumulation over 8 steps. The model is trained for 10 epochs using a learning rate of 1e-4. We adopt a cosine learning rate scheduler with a warm-up ratio of 0.1.
% For VideoLLaMA, we employ a learning rate of 1e-5 for both the large language model (LLM) and multimodal projector, and a lower learning rate of 2e-6 for the vision encoder. The weight decay is set at 0.01. Training uses a gradient accumulation of 4 steps and gradient checkpointing to manage memory efficiently. The cosine learning rate scheduler is also applied here with a warm-up ratio of 0.03, and progress is logged every 5 steps.
% Both setups are optimized for robust and efficient training, ensuring stable convergence and effective learning of fine-grained video understanding tasks.

\begin{table*}
  \centering
  \caption{Comparison of baselines and ours on video captioning and VQA. Subscripts $i$ and $e$ denote the metrics computed on the internal and external splits of the WTS dataset, respectively.}
  \label{tab:cvpr_results}
  \setlength{\tabcolsep}{4pt}      % tighten horizontal padding
  \renewcommand{\arraystretch}{1.0} % default row height

  \begin{adjustbox}{max width=\textwidth}
  \begin{tabular}{lcc*{8}{c}cc}
    \toprule
    \multirow{2}{*}{Model}
      & \multicolumn{8}{c}{Captioning Metrics}
      & \multicolumn{1}{c}{VQA Metric}
      & \multicolumn{1}{c}{Overall Metric} \\
    \cmidrule(lr){2-9} \cmidrule(lr){10-10} \cmidrule(lr){11-11}
      & BLEU-4$_i$$\uparrow$ & METEOR$_i$$\uparrow$ & ROUGE-L$_i$$\uparrow$ & CIDEr$_i$$\uparrow$
      & BLEU-4$_e$$\uparrow$ & METEOR$_e$$\uparrow$ & ROUGE-L$_e$$\uparrow$ & CIDEr$_e$$\uparrow$
      & Acc$\uparrow$ & S2         \\
    \midrule
    VideoLLaMA3-7B \cite{Zhang2025VideoLLaMA3F}
      & 0.2569 & 0.4528 & 0.4512 & 1.1001 
      & 0.2814 & 0.4844 & 0.4658 & 1.2579 
      & 58.6121 & 44.7329 \\
      
    Qwen2.5-VL-7B \cite{Bai2023QwenVLAV}
      & 0.1921 & 0.3968 & 0.3915 & 0.5769 
      & 0.1475 & 0.3483 & 0.3414 & 0.6304 
      & 60.7980 & 42.5136 \\
      
    LLaVA-NeXT-Video-7B \cite{zhang2024llavanextvideo}
      & 0.2412 & 0.4458 & 0.4415 & 1.0122 
      & 0.2603 & 0.4683 & 0.4441 & 1.1276 
      & 49.3216 & 40.3807 \\

    Video-LLaVA-7B \cite{lin2023video}
      & 0.1780 & 0.3553 & 0.3548 & 0.7606 
      & 0.1642 & 0.3201 & 0.3184 & 0.7721 
      & 47.0146 & 35.0327 \\

    MiniCPM-V-2\_6 \cite{yao2024minicpm}
      & 0.2290 & 0.4417 & 0.4352 & 0.8504 
      & 0.2458 & 0.4534 & 0.4429 & 0.9656 
      & 51.3412 & 40.8556 \\

     InternVL3-8B \cite{chen2024internvl}
      & 0.1918 & 0.3951 & 0.3812 & 0.5753 
      & 0.1469 & 0.3475 & 0.3399 & 0.6237
      & 59.9310 & 41.9799 \\
    % add more variants here if needed, e.g. both-only, neither, etc.
    \midrule
    \textbf{VideoLLaMA3-7B \& Qwen2.5-VL-7B (Ours)}
      & \textbf{0.2569} & \textbf{0.4528} & \textbf{0.4512} & \textbf{1.1001}
      & \textbf{0.2814} & \textbf{0.4844} & \textbf{0.4658} & \textbf{1.2579}
      & \textbf{60.7980} & \textbf{45.7572} \\
    \bottomrule
  \end{tabular}
  \end{adjustbox}
\end{table*}

\subsection{Implementation Details}
We implement our experiments using two state-of-the-art Video-Large Language Models: Qwen2.5-VL and VideoLLaMA3.

\noindent\textbf{Qwen2.5-VL.}  
Qwen2.5-VL integrates a dynamic‐resolution Vision Transformer backbone that processes frames at varying sizes without costly normalization, paired with a specialized multimodal projector to align visual tokens with the language embedding space \cite{Bai2023QwenVLAV}.  It achieves precise object grounding via both bounding‐box and point‐based outputs, and handles long videos using absolute time encodings and dynamic frame sampling. We train this model with a batch size of 4, gradient accumulation over 8 steps for 10 epochs at a learning rate of \(1\times10^{-4}\), using a cosine scheduler with a 0.1 warm-up ratio.

\noindent\textbf{VideoLLaMA3.}  
VideoLLaMA3 couples a pretrained vision encoder and a LLaMA‐based language model through a lightweight adaptor, accepting variable‐resolution inputs and merging similar frames at inference to reduce token counts \cite{Zhang2025VideoLLaMA3F}.  Its core joint vision–language alignment module is first co-trained on large-scale image-text corpora, then fine-tuned in a multi-task stage with both image-text and video-text supervision, followed by a video-centric refinement that enhances temporal consistency and motion reasoning.  Training uses learning rates of \(1\times10^{-5}\) for the LLM and projector and \(2\times10^{-6}\) for the vision encoder, weight decay 0.01, gradient accumulation of 4 steps, gradient checkpointing, and a cosine scheduler with a 0.03 warm-up ratio, logging every 5 steps. 
% This complementary approach combines Qwen2.5-VL's precise visual understanding with VideoLLaMA3's coherent temporal narrative generation to provide a comprehensive analysis of traffic scenes.

% \begin{figure*}[h]
%   \centering
%   % Adjust width (or height) as needed:
%   \includegraphics[width=0.8\textwidth]{ICCV2025-Author-Kit/sec/predictions-v2.pdf}
%   \caption{Qualitative comparison of model outputs for traffic safety description. Given input video frames of a pedestrian-vehicle scenario, we compare ground truth annotations with predictions from VideoLLaMA3-7B and Qwen2.5-VL.}
%   \label{fig:my-figure}
% \end{figure*}

\section{Results and Discussion}
Table~\ref{tab:cvpr_results} presents our experimental results comparing the proposed dual-model approach against state-of-the-art baselines. Our integrated VideoLLaMA3-7B and Qwen2.5-VL-7B framework achieves an S2 score of 45.7572, significantly outperforming all individual baselines including LLaVA-NeXT-Video-7B (40.38), InternVL3-8B (41.98), and the best single model VideoLLaMA3-7B (44.73). This 1.09-point improvement over the strongest baseline validates our architectural complementarity hypothesis. Our approach ranks 10th in the Track 2 of the AI City Competition as shown in Table \ref{tab:challenge_top10}.

\begin{table}[h]
\centering
\caption{Top 10 Teams and Their Scores}
\begin{tabular}{|c|l|c|}
\hline
\textbf{Rank} & \textbf{Team Name} & \textbf{Score} \\
\hline
1  & CHTTLIOT                               & 60.0393 \\
2  & SCU\_Anastasiu                         & 59.1184 \\
3  & Metropolis\_Video\_Intelligence        & 58.8483 \\
4  & ARV                                    & 57.9138 \\
5  & Rutgers ECE MM                         & 57.4658 \\
6  & VNPT\_AI                               & 57.1133 \\
7  & AIO\_GENAI4E                           & 55.6550 \\
8  & GenAI4E\_BunBo                         & 52.4267 \\
9  & Tyche                                  & 52.1481 \\
10 & \textbf{MIZSU (ours)}                                 & \textbf{45.7572} \\
\hline
\end{tabular}
\label{tab:challenge_top10}
\end{table}

\begin{figure*}[h]
  \centering
  % Adjust width (or height) as needed:
  \includegraphics[width=0.8\textwidth]{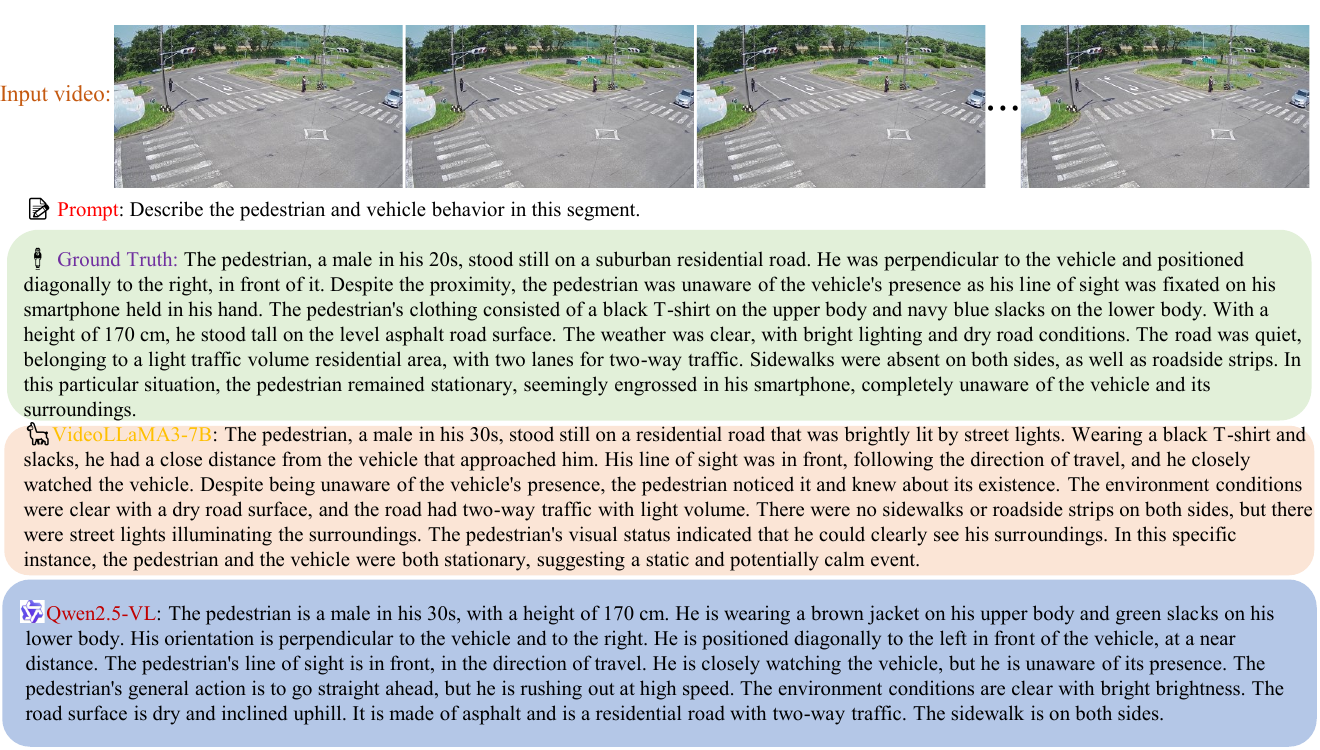}
  \caption{Qualitative comparison of model outputs for traffic safety description. Given input video frames of a pedestrian-vehicle scenario, we compare ground truth annotations with predictions from VideoLLaMA3-7B and Qwen2.5-VL.}
  \label{fig:my-figure}
\end{figure*}

The performance gain stems from utilizing each model's specialized strengths. VideoLLaMA3-7B demonstrates superior captioning with the highest scores across all metrics (BLEU-4: 0.2569, METEOR: 0.4528, ROUGE-L: 0.4512, CIDEr: 1.1001), while Qwen2.5-VL-7B excels at visual reasoning with 60.80\% VQA accuracy. Individual baselines struggle to excel at both tasks simultaneously. For example, LLaVA-NeXT-Video performs well in captioning but achieves only 49.32\% VQA accuracy, while InternVL3-8B shows strong VQA performance (59.93\%) but weaker captioning metrics.

Figure~\ref{fig:my-figure} illustrates why this complementarity matters. In a safety-critical scenario where a pedestrian uses a smartphone while standing still, VideoLLaMA3-7B accurately captures visual details (black T-shirt, stationary behavior) and generates coherent narratives. Conversely, Qwen2.5-VL produces significant hallucinations, misidentifying clothing as "brown jacket and green slacks" and behavior as "rushing out at high speed." These qualitative differences demonstrate that VideoLLaMA3-7B provides essential visual grounding inherent in its architecture for safety descriptions, while Qwen2.5-VL contributes structured visual reasoning capabilities, creating a robust approach that surpasses individual model limitations.

\subsection{Ablation Study}

We conduct an ablation study to validate our training strategy by examining the effects of training on captioning data alone, combined captioning and VQA datasets, and our proposed approach of training separately on captioning and VQA tasks. Table~\ref{tab:ablation_study} summarizes our findings clearly.

\noindent\textbf{Training Only on Captioning Dataset.} When models are trained only on captioning data, we observe high captioning metrics but notably reduced VQA accuracy. For example, VideoLLaMA3-7B attains a CIDEr score of 1.0625 but a low VQA accuracy of 47.50\%. Qwen2.5-VL-7B similarly exhibits CIDEr: 0.5500 and VQA accuracy: 49.00\%. This indicates exclusive caption training limits visual reasoning capabilities.

\noindent\textbf{Training on Combined Captioning and VQA Datasets.} Joint training moderately improves visual reasoning but slightly reduces captioning quality due to task interference. VideoLLaMA3-7B achieves CIDEr: 1.0350 and VQA accuracy: 53.50\%. Likewise, Qwen2.5-VL-7B demonstrates improved visual reasoning (VQA accuracy: 57.50\%) with reduced captioning performance (CIDEr: 0.5450).

\noindent\textbf{Training on Separate Captioning and VQA Datasets (Our Approach).} Our proposed approach which involves training separately on captioning and VQA datasets clearly outperforms other strategies. VideoLLaMA3-7B achieves optimal captioning scores (CIDEr: 1.1001) and strong VQA accuracy (58.6121\%). Qwen2.5-VL-7B records the highest VQA accuracy (60.7980\%) alongside good captioning performance (CIDEr: 0.5769). This strategy effectively addresses task interference and enhances and complementary interplay performance.

\begin{table}[ht!]
  \centering
  \caption{Ablation results highlighting the training strategies on captioning and VQA tasks.}
  \label{tab:ablation_study}
  \setlength{\tabcolsep}{4pt}
  \renewcommand{\arraystretch}{1.1}
  \begin{adjustbox}{max width=\linewidth}
    \begin{tabular}{lcccc}
      \toprule
      Training Method                          & Model                   & BLEU-4$\uparrow$ & CIDEr$\uparrow$ & VQA Acc$\uparrow$ \\
      \midrule
      \multirow{2}{*}{Captioning Only}         & VideoLLaMA3-7B          & 0.2500           & 1.0625          & 47.50             \\
                                               & Qwen2.5-VL-7B           & 0.1850           & 0.5500          & 49.00             \\
      \midrule
      \multirow{2}{*}{Combined Tasks}          & VideoLLaMA3-7B          & 0.2420           & 1.0350          & 53.50             \\
                                               & Qwen2.5-VL-7B           & 0.1830           & 0.5450          & 57.50             \\
      \midrule
      \multirow{2}{*}{\textbf{Separate Tasks (Ours)}} 
                                               & \textbf{VideoLLaMA3-7B} & \textbf{0.2569}  & \textbf{1.1001} & \textbf{58.6121}  \\
                                               & \textbf{Qwen2.5-VL-7B}  & \textbf{0.1921}  & \textbf{0.5769} & \textbf{60.7980}  \\
      \bottomrule
    \end{tabular}
  \end{adjustbox}
\end{table}

% \vspace{-0.8em}

\section{Conclusion}
We presented a dual-model framework that utilizes the complementary strengths of VideoLLaMA3-7B and Qwen2.5-VL-7B for traffic safety analysis. By employing task-specific training, our approach achieves an S2 score of 45.7572, outperforming single-model baselines by 1.09 points. This improvement stems from combining VideoLLaMA's temporal reasoning capabilities with Qwen-VL's superior visual understanding, addressing the inherent trade-off between caption generation and VQA accuracy. The success of our method demonstrates that architectural complementarity, when properly optimized, can overcome the limitations of individual models in safety-critical applications. Future work will explore model distillation to maintain this performance while reducing computational requirements for real-world deployment.
{
    \small
    \bibliographystyle{ieeenat_fullname}
    \bibliography{main}
}


\end{document}